\documentclass{article}
\usepackage{times}

\usepackage{amsmath,amsfonts,bm}

\def\eqref#1{equation~\ref{#1}}

\def\1{\bm{1}}

\DeclareMathAlphabet{\mathsfit}{\encodingdefault}{\sfdefault}{m}{sl}
\SetMathAlphabet{\mathsfit}{bold}{\encodingdefault}{\sfdefault}{bx}{n}

\usepackage{color}
\usepackage{hyperref}
\usepackage{url}
\usepackage{bm}
\usepackage{wrapfig}
\usepackage{ctable}
\usepackage{booktabs}
\usepackage{array,ragged2e}
\usepackage{graphicx}
\usepackage{subfig}
\usepackage{float}

\usepackage{hyperref}

\newcommand\blfootnote[1]{%
  \begingroup
  \renewcommand\thefootnote{}\footnote{#1}%
  \addtocounter{footnote}{-1}%
  \endgroup
}

\usepackage[accepted]{icml2019}

\icmltitlerunning{Understanding and Controlling Memory in Recurrent Neural Networks}

\begin{document}

\twocolumn[
\icmltitle{Understanding and Controlling Memory in Recurrent Neural Networks}



\icmlsetsymbol{equal}{*}

\begin{icmlauthorlist}
\icmlauthor{Doron Haviv}{EE,NBRL}
\icmlauthor{Alexnader Rivkind}{NBRL,MED,WIS}
\icmlauthor{Omri Barak}{NBRL,MED}
\end{icmlauthorlist}

\icmlaffiliation{EE}{Faculty of Electrical Engineering, Technion, Israel Institute of Technology}
\icmlaffiliation{MED}{Rappaport Faculty of Medicine, Technion, Israel Institute of Technology}
\icmlaffiliation{NBRL}{Network Biology Research Laboratory, Technion, Israel Institute of Technology}
\icmlaffiliation{WIS}{Currently at Weizmann Institute of Science, Israel }



\icmlcorrespondingauthor{Doron Haviv}{Doron.Haviv12@gmail.com}
\icmlcorrespondingauthor{Alexnader Rivkind}{Sashkarivkind@gmail.com}
\icmlcorrespondingauthor{Omri Barak}{Omri.Barak@gmail.com}
\icmlkeywords{Recurrent Neural Networks, Dynamics}

\vskip 0.3in]



\printAffiliationsAndNotice{} 

\begin{abstract}

	To be effective in sequential data processing, Recurrent Neural Networks (RNNs) are required to keep track of past events by creating \textit{memories}. 
	While the relation between memories and  the network's hidden state dynamics was established over the last decade, previous works in this direction were of a predominantly descriptive nature focusing mainly on locating the dynamical objects of interest.
	In particular, it remained unclear how dynamical observables affect the performance, how they form and whether they can be manipulated.
	Here, we utilize different training protocols, datasets and architectures to obtain a range of networks solving a delayed classification task with similar performance, alongside substantial differences in their ability to extrapolate for longer delays.
	We analyze the dynamics of the network's hidden state, and uncover the reasons for this difference. Each memory is found to be associated with a nearly steady state of the dynamics which we refer to as a 'slow point'. Slow point speeds predict extrapolation performance across all datasets, protocols and architectures tested. Furthermore, by tracking the formation of the slow points we are able to understand the origin of differences between training protocols. Finally, we propose a novel regularization technique that is based on the relation between hidden state speeds and memory longevity. Our technique manipulates these speeds, thereby leading to a dramatic improvement in memory robustness over time, and could pave the way for a new class of regularization methods.

\end{abstract}

\section{Introduction}  

\blfootnote{Code available at: https://github.com/sashkarivkind/MemoryRNN}

Recurrent Neural Networks (RNN) are the key tool currently used in machine learning when dealing with sequential data \citep{Seq2Seq}, and in many tasks requiring a memory of past events \citep{MINECRAFT}. This is due to the dependency of the network on its past states, and through them on the entire input history. This ability comes with a cost - RNNs are known to be hard to train \citep{RNNDiff}. This difficulty is commonly associated with the vanishing gradient that appears when trying to propagate errors over long times \citep{VanishingGrad}. When training is successful, the network's hidden state represents these memories. Understanding how such representation forms throughout training can open new avenues for improving learning of memory-related tasks.

Linking hidden state dynamics with task-related memories requires some form of reverse engineering. This can be done by focusing on individual recurrent units \citep{KarpahyRNN,MINECRAFT}, or by analyzing global network properties. We opt for the latter, analyzing the RNN's hidden states as a discrete-time dynamical system.

In this framework, memories might be associated with a wide range of dynamical objects. On one extreme, transient dynamics can be harnessed for memory operations \cite{EchoState,MaassLSM,maass2002real}. On the other extreme, there are memory networks \cite{FacebookMemoryNetwork} that memorize \textit{everything} and later use only the relevant memories while ignoring all the rest. The idealized dynamical scenario where each memory is associated with a fixed point in the RNN state space \citep{hopfield1982neural,Sussillo2014NeuralCA,Barak2017RecurrentNN,AttractorsMemory} was refined in \citep{durstewitz2003self,BlackBoxOmri,Mante2013ContextdependentCB} where points which are only approximately fixed (slow points), with a drift that is slower than the task duration were shown to represent memory.

To allow in-depth analysis of the formation of memories throughout training, we analyze a simple delayed classification task. While simple enough to analyze, the task requires both memory and processing - two key operations in many RNN tasks. We train this task using different datasets, unit-types and training protocols, to obtain a range of networks. We find that different protocols lead to comparable performance on the task. A more careful analysis, however, reveals that the resulting networks differ in their extrapolation abilities and reflect their training histories. 

To uncover the underlying reasons for such
 differences, we extend tools used in continuous-time systems in neuroscience \citep{BlackBoxOmri}. We find that memories of the different classes are represented by slow points of varying slowness. We show that the speed of the points predicts the extrapolation properties of their associated class. We establish such a correlation in a large variety of settings.

Having established the importance of slow points as a predictor, we obtain an instructive insight on \emph{how} they evolve along the training course. Detailed analysis of individual training trajectories makes it possible to monitor the formation of slow points under a specific training protocol. This technique uncovers an interplay between newly recruited and functional slow points -- decreasing the stability of the latter in a systematic manner. This provides a link between training curriculum, dynamical objects, memory and performance.

Ultimately, we take a step from merely predicting performance to improving it. To this extent we modify the loss function to penalize  hidden state speed in relevant points, and report a dramatic improvement for extremely long delays.

\section{Task Definition}

\begin{figure*}[h]
\begin{center}
\includegraphics[width=\textwidth]{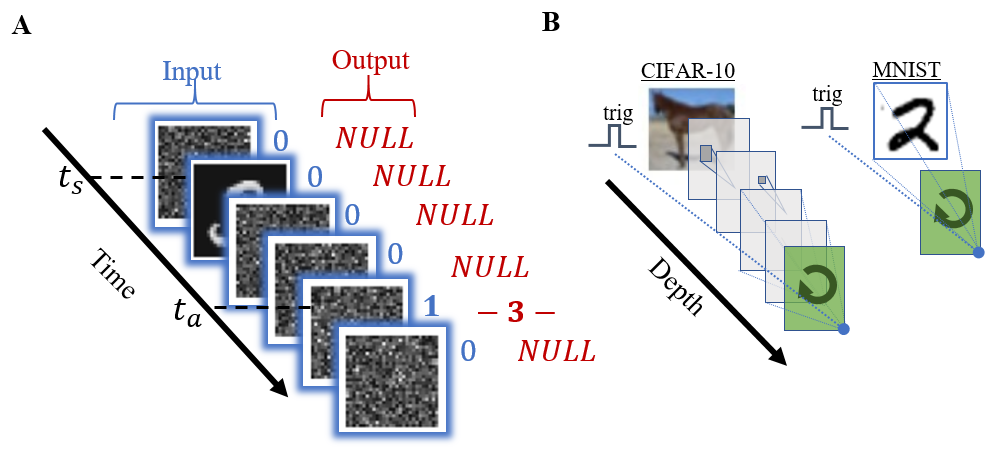}
\end{center}
\caption{\textbf{A} Task. The network is presented an MNIST or CIFAR-10 image amidst noisy images and has to report its label at a later time, as requested by a separate input (${0,1}$ to the right of images). Output should be null at all times except the reporting time. The precise times $t_a, t_s$ vary from trial to trial. \textbf{B} Architecture. In the case of MNIST dataset, both the image and the trigger signal are fed directly into the recurrent layer. For the CIFAR-10 task, a convolutional feed forward network is added in front of the recurrent layer, while the trigger signal is connected directly to the RNN. }
\label{fig:task}
\end{figure*}

Inspired by real-world applications of Recurrent Neural Networks \citep{MINECRAFT}, we designed a task where  the RNN has to combine stimulus processing and memorization (Figure \ref{fig:task}). The network is presented with a series of noisy images, among which appears a single target image (from MNIST or CIFAR-10) at time $t_s$. At a later time point, $t_a$, the network receives a response trigger in a separate input channel, prompting it to output the label of the image. At all other times, the network should report a null label. 

The stimulus and reporting times are chosen randomly at each trial from a uniform distribution on $[1,T_{max}]$ subject to the constraint $t_a-t_s > 4$. The total stimulation time is $T_{max} = 20$, and the network was requested to distinguish between $|V|=10$ different classes of MNIST \citep{MNIST} or CIFAR-10 \citep{CIFAR10}.

Each pixel of the noise mask was sampled from a Gaussian distribution with mean and variance matching its counterpart at the image corpus $\epsilon \sim N(\mu_{n},\sigma^{2}_{n}$). We tested the RNNs ability to extrapolate from this task to longer durations by increasing the delay $T_{max}$.


The motivation behind this task is three-fold. First, as explained, this task is comparable to real-world scenarios where RNNs are used, combining the need for both stimulus memorization and feature extraction. Second, the task lends itself to parametric variations, allowing to compare both different training protocols and generalization abilities. Third, desiring to understand the dynamical nature of memorization in discrete Gated-RNNs, the delay between stimulus and response trigger allows for evolution of RNN hidden-state (HS), which can be reliably analyzed using well known methods from dynamical systems \citep{BlackBoxOmri}, which we modify to our discrete setting.
 
\section{Model}

For MNIST, the network consists of a single recurrent layer of $d=200$ gated recurrent units, an output layer of $|V|+1=11$ neurons, $|V|=10$ neuron for the different classes, and an additional neuron for the null indicator. The input layer has $n+1$ neurons, where $n$ is the number of pixels in the image and an extra binary input channel for the response trigger $X_{r}(t)$ defined by:
\begin{equation}
  X_{r}(t)=\begin{cases}
    1, & \text{if $t=t_{a}$}.\\
    0, & \text{otherwise}.
  \end{cases}
\end{equation}

For CIFAR-10, the network was expanded to $d=400$ recurrent units, along with a convolutional front-end composed of three convolutional layers and two dense layers. To eliminate issues of translational invariance regarding the response trigger and the convolutional front-end, the trigger was added as an extra channel to the final dense layer, right before the recurrent units (Figure \ref{fig:task}).

The gated units are either GRU:

\begin{equation} \label{GRU}
	\begin{array}{l}
    	z = \sigma(W_{z}I+U_{z}h_{t}+b_{z}) \\
    	r =  \sigma(W_{r}I+U_{r}h_{t}+b_{r}) \\
    	h_{t+1} = z\circ h_{t}+(1-z)\circ \tanh(W_{h}I+U_{h}(r\circ h_{t})+b_{h}) \\
	\end{array}
\end{equation}

or LSTM:

\begin{equation} \label{LSTM}
  \begin{array}{l}
  f = \sigma(W_{f}I+U_{f}h_{t}+b_{z}) \\
  i =  \sigma(W_{i}I+U_{i}h_{t}+b_{i}) \\
  o =  \sigma(W_{o}I+U_{o}h_{t}+b_{o}) \\
  c_{t+1} = f\circ c_{t} + i\circ \tanh(W_{c}I+U_{c}h_{t}+b_{c}) \\
  h_{t+1} = o \circ \tanh(c_{t})
  \end{array}
\end{equation}

For the analysis of the network's phase space, we denote the state of the recurrent layer by $\xi$, which for LSTM is $\xi=\begin{pmatrix} h \\ \tanh(c) \end{pmatrix}$ and for GRU $\xi = h$. 

The network was trained using the 'Adam' optimizer \citep{ADAMOptim} with a soft-max cross-entropy loss function with an increased loss on reporting at $t=t_{a}$ in proportion to $T_{max}$. Full description of each protocol, including schedules and other hyper-parameters is given in the supplemental code.

\section{Training Protocol: Two Types of Curricula}
We found that training failed when using straightforward stochastic gradient (SG) optimization on the full task. The network converged to a state where it consistently reports 'null' without regarding neither the output trigger nor the images it has received as inputs. This suboptimal behavior did not improve upon further training.
On the other hand, we observed that simpler versions of the task are learn-able. If the maximal delay between stimulus and reporting time was short or when we introduced only a limited number of different digits, the network was able to perform the task. This led us to try two different protocols of curriculum learning \citep{CurrLearning} in order teach the network the full required task:

1. Vocabulary curriculum (VoCu) - here we started from two classes $V=\lbrace c_1,c_2 \rbrace$  and then increased the vocabulary gradually until reaching the full  class capacity. This protocol is similar to the original concept of \citep{CurrLearning} except the fact that in our vocabulary all the classes occur with the same frequency, and the selected order of class introduction is in fact arbitrary. 

2. Delay curriculum (DeCu) - starting from short delays between stimulus and reporting times ($T_{max}=6$), we progressively extended it toward the desired values. Implicitly mentioned in \citep{VanishingGrad}, this regime is expected to mitigate the vanishing gradient problem, at least during initial phase of training.

\section{Extrapolation Ability Depends on Training Protocol}
\label{sec:GenAbility}
We found that,  in good accordance with existing literature \citep{CurrLearning,RNNArch}  results for the nominal test-set were fairly indifferent to the training protocol (Supplementary material). Once we evaluated the ability of each setting to extrapolate to longer delays, however, similarity ends and differences emerge.


\begin{figure}[h]
\begin{center}
\includegraphics[width=0.45\textwidth]{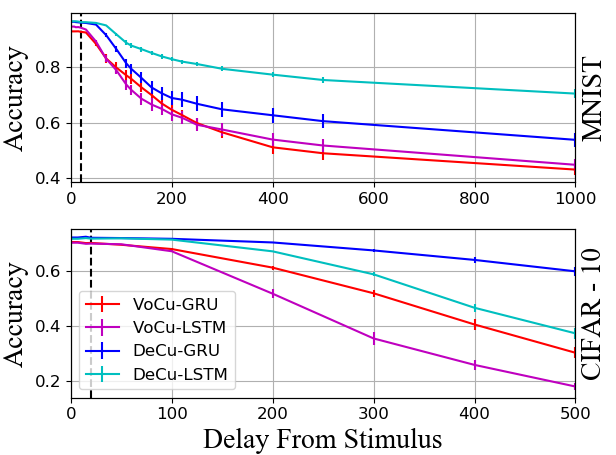}
\end{center}
\caption{Retrieval accuracy when increasing the delay between stimulus and response trigger beyond $T_{max}=20$. Despite similar performance initially, the ability to generalize for greater delays than those trained for (dashed vertical lines) varies with protocol. DeCu was superior to Vocu in both LSTM and GRU architectures for both MNIST and CIFAR-10 datasets.}
\label{fig:ExpDiff}
\end{figure}

We observed how each setting performs when the delay between stimulus and response trigger is extended further beyond $T_{max}=20$. If $|V|=10$ robust fixed-point attractors have formed, retrieval accuracy should not be affected by the growing delay. If the computation is based on transients, then all class information is expected to eventually vanish.
 
Experiments revealed that neither of these extreme options was the case - performance deteriorated with increasing delay, but did not reach chance levels  (Figure \ref{fig:ExpDiff}). This deterioration implies that not every memorized digit corresponds to a stable fixed point attractor, but some do. Furthermore, the deterioration was curriculum-dependent, with DeCu outperforming VoCu in all cases. 





\section{Dynamics of Hidden Representation} 
\label{sec:dynamics}

\begin{figure}[h]
\begin{center}
\includegraphics[width=0.45\textwidth]{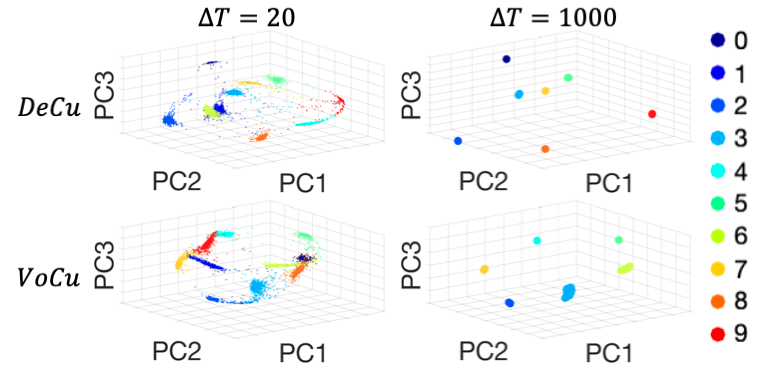}
\end{center}
\caption{Hidden state projected on leading principal components in GRU - RNN  on MNIST for delays of $\Delta t= 20 \; \text{and} \; 10^{3}$ time-steps from stimulus. States are color codded by their prediction. For the nominal delay ten distinct regions are observed in the state space, corresponding to each of the  $|V|=10$ classes. Examination of a larger delay $\Delta t =10^3$ reveals that some clouds collapse into a single point at the center of the cloud, while others vanish completely. The spread of samples in the nominal delay, along with a smaller number of distinct fixed points at $\Delta t=10^{3}$ in VoCu aligns with our findings of faster and less stable dynamics compared to DeCu.}
\label{fig:HiddenStatePC}
\end{figure}

\begin{figure}[h]
\begin{center}
\includegraphics[width=0.45\textwidth]{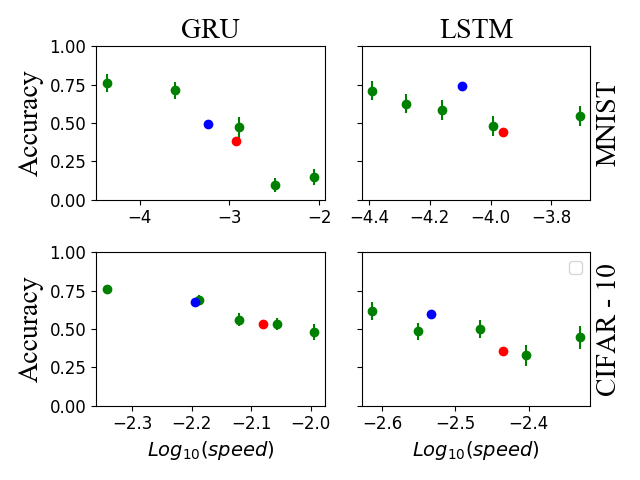}
\end{center}
\caption{Extrapolation accuracy predicted by slow point speeds. The accuracy at long delays ($\Delta t =10^3$ for MNIST and $\Delta t =500$ for CIFAR-10) is shown as a function of the slow point speed of the associated class (green, errorbars denote standard error of the mean). In all datasets, unit-types and training methods, slower speeds correlate with increased accuracy. The red and blue dots show the mean speed and accuracy for each training protocol (standard error of the mean is smaller than marker size). The difference in speeds between the protocols underlies the different extrapolation performance shown in Figure \ref{fig:ExpDiff}. Ten networks were used for MNIST, and five for CIFAR-10.} 
\label{fig:SpeedStem}
\end{figure}

The relevant phase space of this dynamical system is the recurrent layer state $\xi$. We thus begin by visually inspecting (in the first 3 PCA components) the activity of the network for the maximal training delay, $\Delta t=20$. We show here results for the MNIST dataset with a GRU architecture, but similar behavior is seen for other conditions and the statistics of all conditions is analyzed below. The left panels of Figure \ref{fig:HiddenStatePC} show that different trials of each digit are well separated into $|V|$ regions with a one to one correspondence to data classes. Following these trajectories for a longer delay of $\Delta t=1000$ shows that some regions converge into what appears to be fixed points, while others vanish (right panels). These figures also clearly show the difference between the two protocols. While both achieve a good separation with the nominal delay (left), it is already apparent that VoCu leads to clouds of points with a larger spread, possibly indicating a weaker attraction. 

To verify the existence or absence of fixed points hinted by the above visualization, we apply an algorithm developed for continuous time vanilla RNN \cite{BlackBoxOmri} to our setting. Briefly, fixed points (stable or unstable) are local minima of the (scalar) speed $S(\xi,I)$ of the hidden state $\xi$. 

\begin{equation}
S(\xi,I)=||F(\xi,I)-\xi ||_2
\label{eq:speed}
\end{equation}
where the evolution of state,
\begin{equation}
F(\xi')=\xi(t+1) \bigg \vert _{\xi(t)=\xi'} 
\end{equation} 
is given by equations \ref{GRU} or \ref{LSTM} for GRU or LSTM respectively. 
It is now possible to use gradient descent on the speed $S$ with respect to state $\xi$, namely, $\nabla_\xi S$, to locate such minima.


The initial conditions for this gradient descent were obtained by running the network with the mean delay value $\Delta t = 15$, and using the averages of hidden states of each class as initial conditions. The external input $I$ during gradient descent was the average of the noise images, thus effectively making our system Time-Invariant so that such points and their stability are well defined. We verified that using different fixed external inputs did not qualitatively alter the results (not shown). We repeated the procedure for several realizations, and it always resulted in a local minimum of speed for each class  (which we call a slow point), with a readout that matches the class label.

To look for a \textit{quantitative} relation between slow point speed and the memory robustness we located slow points for every class, computed their speed and the prediction accuracy of their associated classes after a long delay.  Figure \ref{fig:SpeedStem} shows that the speed of the slow point associated with a certain class can predict how members of that class will react to extrapolation experiments. This trend holds for all architectures, unit types and datasets tested. 

The colored dots in Figure \ref{fig:SpeedStem} denote the mean of the speeds obtained by the two different protocols. The picture here is consistent with that observed in Figure \ref{fig:ExpDiff}, with DeCu outperforming VoCu for all cases. Our results suggest that this difference is mediated by a difference in speeds of the associated slow points. 
 
 We also trained networks on an additional task of delayed \emph{matching} (as opposed to classification), and observed the same speed-accuracy anticorrelation. (Supplementary material)

\section{Formation of Slow Points - Why Protocols Differ}
\label{Sec:SlowFormation}
We saw that the two training protocols lead to a different representation of the stimulus memory by the network, and hence to different dynamical objects. How does training give rise to these differences? To answer this question, we analyze in detail two settings - a GRU and LSTM architectures trained on the MNIST database. We follow the slow points of the velocity backwards in training time to learn how they emerge and change throughout training, and correlate these events with network performance.

\begin{figure*}[h!]
   \centering
   \includegraphics[width=1\textwidth]{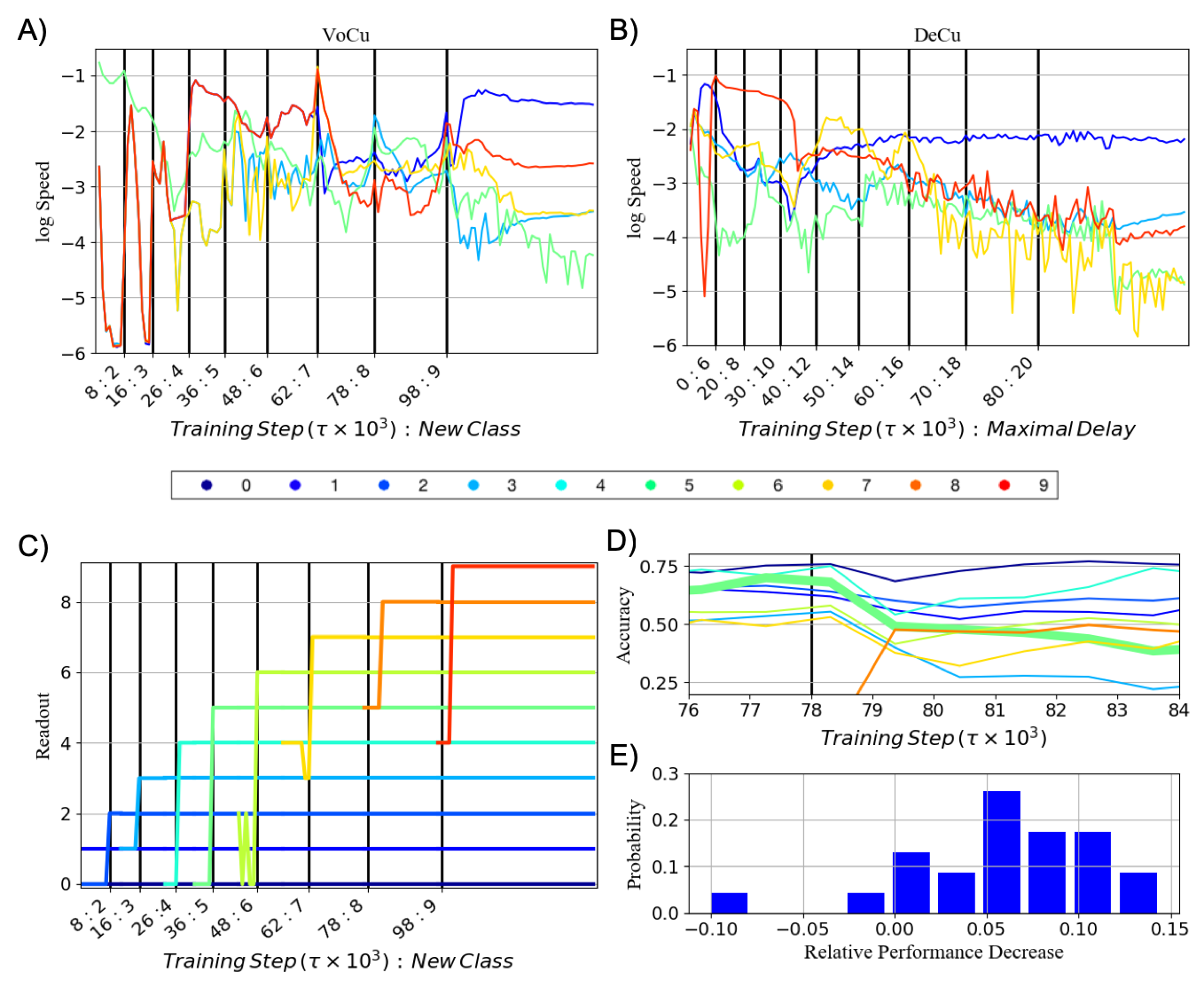} 
   \caption[Caption for LOF]{\textbf{A,B} Speeds of each slow-point through training (only five are shown for clarity) obtained by iteratively tracking them back in training time. The specific schedule of each curriculum is marked on the time axis ($\tau_i : c_i$ for VoCu and $\tau_i : T_{max,i}$ for Decu). VoCu shows sharp jumps in the speed of all points for each class introduction, in contrast to DeCu which exhibits a gradual slowing down along the whole course of training. \textbf{C} A branching diagram for VoCu, obtained from short backtracking every class, proximal to its appearance. The change in readout suggests which slow point gave rise to the new point (e.g., 8 from 5 at time 78; 7 from 3 or 4 at time 62). \textbf{D} The accuracies of each individual class for the same VoCu realization when class '8' was introduced. As a result, all previously existing classes show a degradation in accuracy, however, the class from which class '8' emerged from (class '5') exhibited a stronger decline. Noisy images with double amplitude ($2\sigma_n^2$) were used to amplify the effect strength. \textbf{E} Verifying the statistical significance of the result in (D). For each branching event, we compared the accuracy drop of classes that gave rise to the new class, with the drop for the remaining classes. The histogram is from all events in three VoCu realizations. It shows that indeed accuracy of spawning classes decreases more following a branching event.}
\label{fig:BackTrack}
\end{figure*}

We located slow points as described in Section \ref{sec:dynamics} , and then used them as initial conditions for gradient descent on the network defined by the previous training step. We then repeated this procedure iteratively for all training steps. The assumption is that the change in network parameters at each training step will not induce a very large shift in the locations of the relevant slow points, and thus our continuation procedure can track them. This is not clear in the case of VoCu, where one might expect rapid changes whenever a new class is added. Looking at the speeds of the tracked slow points shows that this is indeed the case for VoCu (Figure \ref{fig:BackTrack}, \textbf{A,B}), with all tracked speeds exhibiting such jumps. DeCu, on the other hand, shows a gradual slowing down of the slow points throughout training.

By observing the gradual slowing down in DeCu, it is easy to understand why this protocol improves performance - slow points become slower. But the situation for VoCu is more complicated because introducing new classes qualitatively changes the dynamics. A natural expectation might be that the classes that were presented first will have more time to stabilize, and thus will be the slowest. We saw that this is not the case (not shown), and thus proceeded to look deeper into the class introduction events along VoCu training.

Consider such an event - the introduction of class $c_i$ at training time $\tau_i$ (We use $\tau$ to denote training step (of SG), as opposed to the time during each trial which is denoted by $t$). We perform a short backtracking procedure - starting just before the succeeding class is introduced ($\tau_{i+1}-2000$), and following it back until slightly before it appeared ($\tau_i-2000$). We verified that we can follow the point despite the jumps mentioned above (shown in supplementary material).

We discovered that the new class is assigned to an existing slow point. This slow point was previously classified as an existing class $j<i$. By stitching together all such backtracking procedures, we obtain a diagram indicating where each new class originated  (Figure \ref{fig:BackTrack}\textbf{C}).

Does this history affect performance? To answer this question, we checked the difference in performance of the various classes following the introduction of a new one. For instance, the diagram shows that class 8 originated from class 5. Figure \ref{fig:BackTrack}\textbf{D} shows that the performance of class 5 was impaired more than other existing classes following the introduction of class 8. 

To evaluate the statistics of this phenomenon, we repeat this procedure for many networks, and all class introduction events. Figure \ref{fig:BackTrack}\textbf{E} shows that the decline in accuracy of the classes that spawned the new class is significantly larger than that of the other classes (LSTM histogram is shown in supplementary material). Specifically, for class $c_i$ being newly introduced at training step $\tau_i$, accuracy of all classes $\lbrace c_j \rbrace_{j<i}$ is evaluated at training steps $\tau=\tau_i-1000$ and $\tau=\tau_i+4000$. Accuracy change $\Delta a_k$ for class $c_k$ that branches into class $i$ is compared to the average $ \left<  \Delta a_j \right> _{j<i,j\ne k}$.

\section{Improving Long Term Memory}
\label{Sec:SpeedReg}
Can the aforementioned insights help improve performance? Can we obtain memory for delays that are substantially longer than in the scenarios used for training? To answer this question, we first trained the network as before using one of the protocols discussed above. We then trained for an additional period of $5000$ gradient steps (compared to $140000$ in initial training) while regularizing our standard cross-entropy loss function $L_{xent}$ by a term accounting for hidden state speed. The new loss: 
\begin{equation}
    L' = L_{xent}+\lambda \sum_{i \in V} |S(\bar{\xi_i})|
\end{equation}
penalizes for high speed $S$ of \eqref{eq:speed} at representative points $\bar{\xi_i}$ associated with each class.

\begin{figure}[h]
\begin{center}
\includegraphics[width=0.45\textwidth]{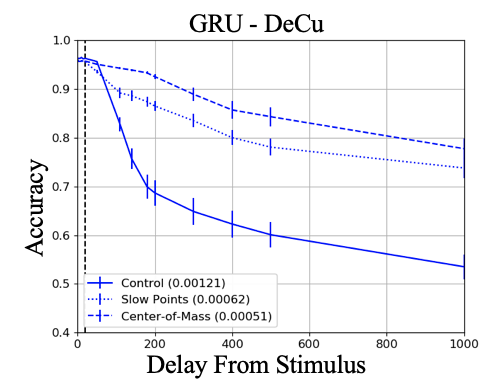}
\end{center}
\caption{Effect of speed regularization on the performance is demonstrated for the DeCu training on GRU architechure on MNIST, results for all other setting are in Supplementary material. For control, we trained the network for the same number of additional gradient decent steps without any regularization (solid). Regularization targets were either the slow-points (dotted) or the center-of-mass (dashed). Both regularization methods resulted in dramatic improvements compared to control, which is also reflected in a smaller speed of slow-points after additional training (shown in legend brackets).}
\label{fig:SpeedReg}
\end{figure}

The natural candidates for $\bar{\xi_i}$ are the slow points discussed above, and indeed Figure \ref{fig:SpeedReg} shows that dramatic improvements were achieved, when compared to the same training without the added regularization. Using the slow points as regularizer targets is still somewhat costly, as their detection requires a gradient descent step. Furthermore, as training proceeds, the location of slow points can move -- rendering the original regularization targets less effective. We thus used a proxy for the slow points by taking $\bar{\xi_i}$ to be the centre of mass of each class. This simpler procedure achieved results that were comparable to using the slow points themselves. In both cases, we verified that slow point speed was manipulated to achieve the improved performance (Figure \ref{fig:SpeedReg} legend), and that the accuracy for nominal delays remained virtually intact. 
In principle, one could train for such long delays by straight forward back propagation through time, but this would be orders of magnitude more time consuming than our method, if not impossible.

\section{Discussion}
Training RNNs to perform memory-related tasks is difficult \citep{pascanu2013difficulty}, and as a consequence many suggestions were made on how to alleviate this difficulty. Changing network architecture, unit-types or training protocols might be expected to generate different solutions to the same task. 



Here we showed that different training protocols can lead to different locally optimal solutions. Although these solutions perform similarly under nominal conditions, challenging the networks with unforeseen settings reveals their differences.

An RNN is a dynamical system, and as such its operation can be understood in the language of fixed points and other dynamical objects. By analyzing the phase space of the network's hidden states, we showed that the memory of each class was associated with a slow point of the dynamics. Such slow points were shown to assist network functionality \citep{machens2005flexible}, and arise through training \cite{Mante2013ContextdependentCB,durstewitz2003self}. The speeds of these slow points were highly correlated to the functional characteristics of memory longevity, and thus provide a dynamical explanation of the idiosyncrasies observed between curricula and architectures. Our result proved valid across architectures, datasets and unit types.

In  specific  cases,  we  were  able  to  follow  the  formation of the recruitment of slow points to representation of memories during the training process. Such recruited slow points reside in an area of phase space that \textit{belongs} to a specific existing class. We showed that this process is associated with a decrease in network accuracy that is specific for the classes that contribute the slow point.
Things could have been different - slow points could emerge in an area of phase space that is distant from existing ones, and the introduction of a new class could have resulted in a uniform effect on all existing classes. To uncover this process, we introduced a back-tracking  methodology  that  could  be  relevant  to  any  case in which learning modifies the dynamics of the network. Possible applications include studying the success and failure in creating memories \citep{DynamicsOfDyanmics}, preventing catastrophic forgetting, understanding memory capacity and more.

The setting studied in this work is a particular case of the more general problem of solution equivalence in gradient based optimization: On the one hand theoretical and numerical evidence does exist for training outcome's indifference to protocol details.  Specifically to our case of RNN, it is shown in \citep{VisCurr} that, at least for language modeling tasks, performance does not heavily depend on training protocol. On the other hand, such an indifference is far from being fully established. In particular, stochastic gradient optimization suffers from known drawbacks \citep{dauphin2014identifying,martens2011learning} and might prove dependent on initialization \citep{sutskever2013importance} (and in particular pre-training). Our results show that networks with the same initialization can reach different solutions, and begin to uncover the dynamics underlying the route to these solutions. 

Ultimately, at the engineering end of this study, the aforementioned insights allowed us to come up with a novel regularization technique which, by penalizing the hidden state speeds, enables a dramatic improvement of performance for extremely long times, while keeping nominal performance intact. Having noticed that slowly increasing the delay (DeCu) resulted in better stability and increased extrapolation ability for all the conditions tested, it might have been possible to reach similar performance by extending the DeCu protocol to very large delays. Such long backpropagation through time, however, is extremely costly, if not impossible, and avoided by our new method. Notably, our method preserves linear time complexity of plain RNN, as opposed to attention mechanisms whose complexity is quadratic in time (\citet{ke2018sparse} and references therein).

Importantly, our regularization procedure relies on an informed guess of \emph{what} information the system needs to memorize to improve performance and of \emph{where} such  information may be stored. In a delayed classification task, the trivial and straight forward guess is that remembering \emph{class} improves  performance with relevant information being represented by the mean of class' samples hidden state $\bar{\xi_i}$.  Future work may generalize our methodology to more complex tasks, such as natural language processing, where long term memorization amid a continuous stream of potentially useful input remains an open challenge \citep{paperno2016lambada}. Doing so will require identifying the appropriate entities to memorize and the locations in hidden space representing these memories  \cite{bau2018identifying}, for use as regularization targets.

\subsection*{Acknowledgements}
 OB is supported by Israel Science Foundation (346/16). The Titan Xp used for this research was donated by the NVIDIA Corporation.

\bibliography{example_paper}
\bibliographystyle{icml2019}


\newpage

\twocolumn[
\icmltitle{Supplemental Material}
]

\appendix
\section{Delayed Template Matching Task}

\begin{figure}[h]
\begin{center}
\includegraphics[width=0.45\textwidth]{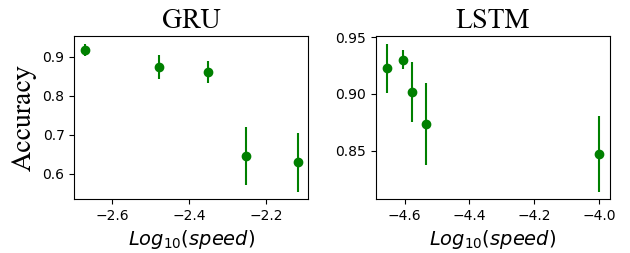}
\end{center}
\caption{Speed of slow-point and the accuracy of the corresponding class for a long delay on the delayed template matching task. Similarly to the main text Figure 4 both architectures exhibit a clear negative correlation between slow-point speed and the effectiveness for long delays. Note that the delayed template matching task was learn-able without any curricula, further generalizing our findings to naive training. Ten networks for each architecture were used.}
\label{fig:SpeedStemTask2}
\end{figure}

 To examine the relevance of our results to other tasks, we considered a delayed template matching setup (also known as delayed match to sample). Here instead of introducing a trigger via an extra input channel, we introduce a second stimulus. When the second stimulus is introduced, the system is to report whether both stimuli belong to the same class or not. At all other times, the system should report a null output. We trained both architectures on the MNIST dataset, with the same minimal and maximal delays as the original task and an equal probability of both stimuli belonging or not belonging to the same class. As a result of the simpler nature of the template matching task compared to the classification task, both architectures were able to perform well without curricula training for nominal delays. We then repeated the analysis reported in Figure 4 of the main text. Figure \ref{fig:SpeedStemTask2} demonstrates here, similarly to the original task, the predictive nature of slow-point speeds. The faster the slow-point is, the less likely is the memory of the corresponding class to sustain for long delays.

\section{LSTM Branching}

\begin{figure}[h]
\begin{center}
\includegraphics[width=0.45\textwidth]{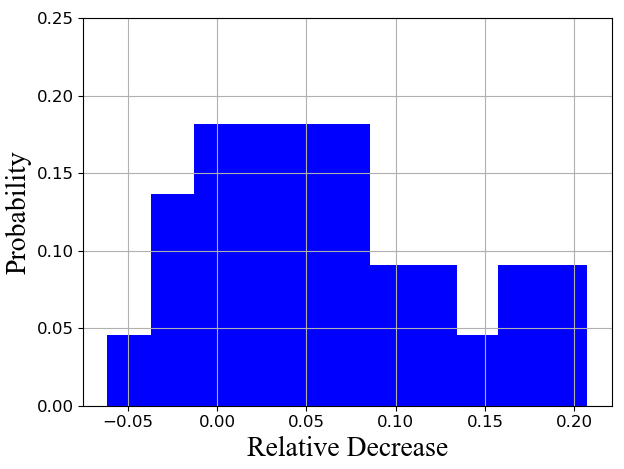}
\end{center}
\caption{Same as main text Figure 5E but for LSTM instead of GRU. Here as well, classes which bifurcated when new classes were spawned were more adversely affected compared to classes which did not experience a bifurcation.}
\label{fig:BifLSTM}
\end{figure}

Performing the Bifurcation analysis of Section 7 but for the LSTM architechure instead of GRU. Just as for GRU, Figure \ref{fig:BifLSTM} shows that the details of slow-point bifurcations affect performance. In the VoCu protocol, the introduction of new classes is accompanied by a bifurcation of an existing slow point. The classes associated with this spawning slow point are significantly adversely affected by this event, compared to other classes.

\section{Regularization Figures}
\label{App:SpeedReg}

\begin{figure*}[h]
\begin{center}
\includegraphics[width=0.95\textwidth]{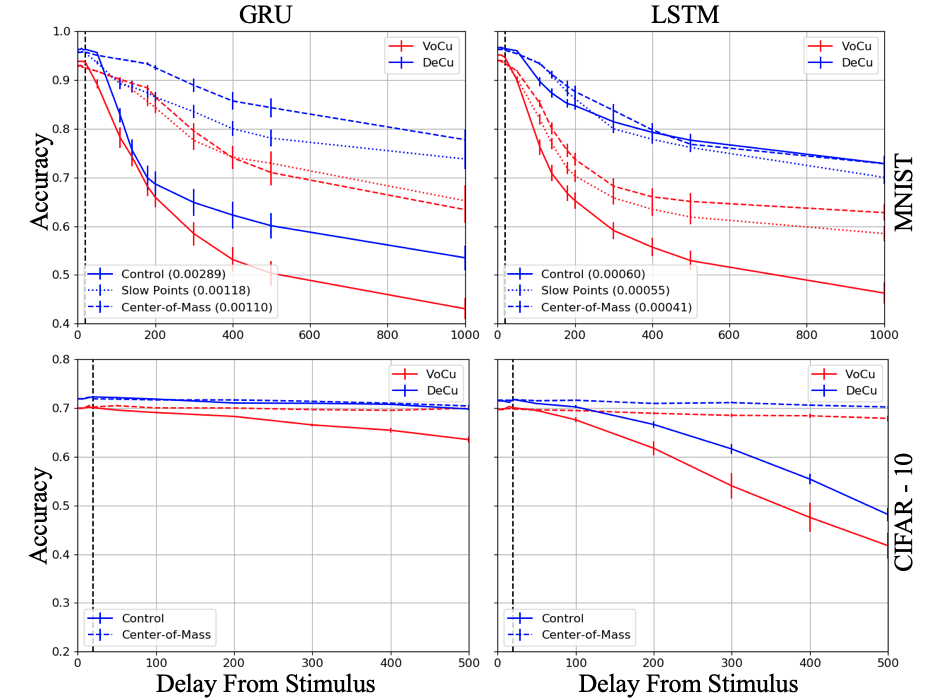}
\end{center}
\caption{Effect of speed regularization on the performance is demonstrated for all the test cases from Figure 6.}
\label{fig:SpeedRegAll}
\end{figure*}

Results of Section 8 to all the data sets, training curricula and unit types are reported in Figure \ref{fig:SpeedRegAll}. Following the trend of section 8, when regularizing the speed of the center-of-mass of each class or of the appropriate slow-point superior results were achieved for extended delays compared to when no regularization was applied ($\lambda \equiv 0 $).

\section{Validity of backtracking procedure during introductions of new classes in VoCu}
\begin{figure*}[h]
\begin{center}
\includegraphics[width=0.95\textwidth]{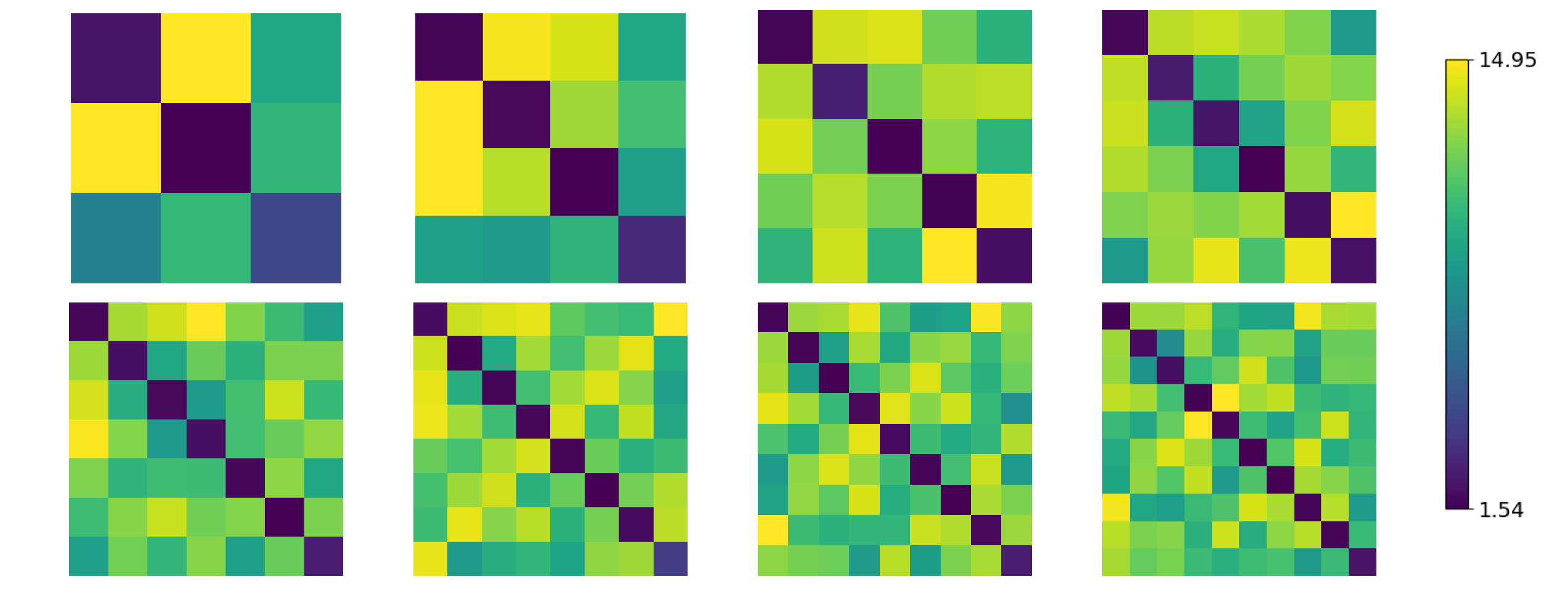}
\end{center}
\caption{Displacement of slow points along the training steps where new classes introduced is shown for all steps of a VoCu training example. Newly introduced class is always at the bottom-right corner.}
\label{fig:DisplacementDuringBackTrack}
\end{figure*}

Backtracking of slow point speeds around an event of introduction of new class reveals a spurious behavior of these speeds. We validated that even at these training epochs the tracking follows specific slow points and does not just capture random slow points in the hidden representation space. 
Fig. \ref{fig:DisplacementDuringBackTrack} depicts displacements $|\xi_k(\tau_i-1)-\xi_k(\tau_i+1)|$ of slow points associated with classes $1\le k \le i$ back tracked along the step of introduction of a new class $c_i$, and compares them to distances to other slow points being tracked. It confirms that displacements of the points claimed to be tracked are indeed systematically lower than distances from other slow points under tracking. Importantly this is also true for the slow point $\xi_i$ which is assigned to a newly introduced class.

\section{Accuracies on Test and Train sets for nominal task} \label{App:AccTables}
Accuracies on the MNIST and CIFAR-10 train and test sets for the nominal task, for both GRU and LSTM architectures and the described curricula are reported in tables \ref{table:MNISTGRU},\ref{table:MNISTLSTM},\ref{table:CIFARGRU},\ref{table:CIFARLSTM}.

\begin{table}[h]
	\caption{MNIST - GRU}
	\label{table:MNISTGRU}
	\begin{center}
		\begin{tabular}{llllll}
			&  &DeCu	&VoCu \\ 
			\hline\hline \\ 
			Training Set	&Null	&100\%		&100\%  \\
			&Digits	&$97.5\pm0.2$\%	&$94.3\pm0.8$\%      \\
			\hline \\
			Test Set	&Null	&100\%	&100\% \\
			&Digits	&$94.3\pm0.8$\% 	&$94.3\pm0.8$\% \\
		\end{tabular}
	\end{center}
\end{table}

\begin{table}[h]
	\caption{MNIST - LSTM}
	\label{table:MNISTLSTM}
	\begin{center}
		\begin{tabular}{llllll}
			&  &DeCu	&VoCu \\ 
			\hline\hline \\ 
			Training Set	&Null	&100\%		&100\%  \\
			&Digits	&$97.3\pm0.1$\% &$95.2\pm0.3$\%      \\
			\hline \\
			Test Set	&Null	&100\%	&100\% \\
			&Digits	&$97.3\pm0.1$\%	&$95.2\pm0.2$\%        \\
		\end{tabular}
	\end{center}
\end{table}

\begin{table}[h]
	\caption{CIFAR - GRU}
	\label{table:CIFARGRU}
	\begin{center}
		\begin{tabular}{llllll}
			&  &DeCu	&VoCu \\ 
			\hline\hline \\ 
			Training Set	&Null	&100\%		&100\%  \\
			&Digits	&$97.12\pm0.2$\%	&$97.0\pm0.7$\%      \\
			\hline \\
			Test Set	&Null	&100\%	&100\% \\
			&Digits	&$72.0\pm0.3$\%		&$69.8\pm0.3$\%       \\
		\end{tabular}
	\end{center}
\end{table}

\begin{table}[h]
	\caption{CIFAR - LSTM}
	\label{table:CIFARLSTM}
	\begin{center}
		\begin{tabular}{llllll}
			&  &DeCu	&VoCu \\ 
			\hline\hline \\ 
			Training Set	&Null	&100\%		&100\%  \\
			&Digits	&$96.8\pm0.2$\%	&$97.1\pm0.2$\%      \\
			\hline \\
			Test Set	&Null	&100\%	&100\% \\
			&Digits	&$71.2\pm0.5$\%		&$69.8\pm0.8$\%       \\
		\end{tabular}
	\end{center}
\end{table}

\end{document}